\documentclass[11pt,a4paper]{article}

\usepackage[utf8]{inputenc}
\usepackage[T1]{fontenc}
\usepackage{mathpazo}          
\usepackage{amsmath,amssymb}
\usepackage{graphicx}
\usepackage{booktabs}
\usepackage{array}
\usepackage{tabularx}
\usepackage[margin=1in]{geometry}
\usepackage{setspace}
\usepackage{natbib}
\usepackage{url}
\usepackage[hidelinks]{hyperref}
\usepackage{xcolor}
\usepackage{float}
\usepackage{enumitem}
\usepackage{fancyhdr}
\usepackage{caption}
\usepackage{tikz}
\usetikzlibrary{arrows.meta,positioning}
\newcommand{\doi}[1]{\href{https://doi.org/#1}{doi:#1}}

\title{\textbf{The Cognitive Divergence:}\\[6pt]
AI Context Windows, Human Attention Decline,\\
and the Delegation Feedback Loop}

\author{Netanel Eliav\\[4pt]
Machine Human Intelligence Lab (MHIL)\\
\texttt{netanel@mhil.org} \quad \texttt{inetanel@me.com}\\
\url{https://mhil.org}}

\date{March 2026\\[6pt]
\small Preprint --- submitted to arXiv cs.AI, cs.HC\\
\small This paper has not undergone peer review.}

\begin{document}
\maketitle
\thispagestyle{fancy}

\begin{abstract}
This paper documents and theorises a self-reinforcing dynamic between two measurable trends: the exponential expansion of large language model (LLM) context windows and the secular contraction of human sustained-attention capacity. We term the resulting asymmetry the \emph{Cognitive Divergence}. AI context windows have grown from 512 tokens in 2017 to 2,000,000 tokens by 2026 (factor $\approx$3,906; fitted $\lambda = 0.59$ yr$^{-1}$; doubling time $\approx 14$ months). Over the same period, human Effective Context Span (ECS)---a token-equivalent measure derived from validated reading-rate meta-analysis \citep{Brysbaert2019} and an empirically motivated Comprehension Scaling Factor---has declined from approximately 16,000 tokens (2004 baseline) to an estimated 1,800 tokens (2026, extrapolated from longitudinal behavioural data ending 2020 \citep{Mark2023}; see Section~9 for uncertainty discussion). The AI-to-human ratio grew from near parity at the ChatGPT launch (November 2022) to 556--1,111$\times$ raw and 56--111$\times$ quality-adjusted, after accounting for retrieval degradation \citep{Liu2023,Chroma2025}. Beyond documenting this divergence, the paper introduces the \emph{Delegation Feedback Loop} hypothesis: as AI capability grows, the cognitive threshold at which humans delegate to AI falls, extending to tasks of negligible demand; the resulting reduction in cognitive practice may further attenuate the capacities already documented as declining \citep{Gerlich2025,KimEtal2026,KosmynaEtal2025}. Neither trend reverses spontaneously. The paper characterises the divergence statistically, reviews neurobiological mechanisms across eight peer-reviewed neuroimaging studies, presents empirical evidence bearing on the delegation threshold, and proposes a research agenda centred on a validated ECS psychometric instrument and longitudinal study of AI-mediated cognitive change.
\end{abstract}

\noindent\textbf{Keywords:} large language models; context window; sustained attention; cognitive offloading; delegation threshold; working memory; cognitive load theory; transformer architecture; prefrontal cortex; dopamine; human--AI interaction; effective context span; deskilling.\\[4pt]
\noindent\textbf{arXiv categories:} cs.AI (primary); cs.HC (cross-listed)\\[8pt]

\newpage
\tableofcontents
\newpage

\section{Introduction}

Consider a routine workplace interaction in 2026: a professional opens an AI assistant and types, \emph{``Write a two-sentence reply declining this meeting.''} The request concerns a task that requires perhaps thirty seconds of unassisted cognitive effort. The model obliges instantly. The professional accepts the output, copies it, and moves on. This transaction is unremarkable. That is precisely the point. It is a representative instance of how many knowledge workers now engage with LLM systems in daily practice \citep{MicrosoftLinkedIn2024,Noy2023}.

This paper argues that this transaction is a symptom of a structural shift in the boundary between human and artificial cognition, one that operates as a self-reinforcing loop. As AI systems grow capable of processing ever-larger and more complex contexts, the cognitive threshold at which humans choose to delegate tasks to those systems falls correspondingly. Tasks once performed with minimal effort are now routinely offloaded. The downstream consequence, supported by emerging empirical evidence, is further attenuation of the very attentional and compositional capacities that the behavioural literature documents as already declining \citep{Mark2023,Gerlich2025,KimEtal2026}.

The argument rests on two independently documented empirical trends. AI context windows have expanded from 512 tokens (the original transformer, \citealt{Vaswani2017}) to 2,000,000 tokens (Grok~4.20, 2026), a factor of approximately 3,906 in nine years. The human trajectory runs in the opposite direction. The most extensive longitudinal study of digital sustained attention documents that mean screen-focus duration among knowledge workers stabilised at approximately 47 seconds by 2016--2020 \citep{Mark2016}, a figure the principal investigator characterises as a new attentional equilibrium rather than a continuing decline \citep{Mark2023}. Translating the human attentional trajectory into tokens via validated reading-rate meta-analysis yields what this paper terms the \emph{Effective Context Span (ECS)}: an estimated decline from approximately 16,000 tokens (2004) to approximately 1,800 tokens (2026 extrapolation). The \emph{Cognitive Divergence} is the measurable, widening gap between these two trajectories.

The Cognitive Divergence has been implicit in several bodies of literature, including AI scaling \citep{Vaswani2017,Brown2020}, attention research \citep{Mark2023,Mark2016}, and cognitive offloading \citep{Gerlich2025,KimEtal2026}, but has not previously been quantified on a common scale, nor has its feedback mechanism been named and formalised. This paper does both, and in doing so establishes a terminological and empirical foundation for subsequent research.

\subsection{Contributions}

This paper makes the following original contributions:

\begin{enumerate}[leftmargin=*]
\item The \textbf{Effective Context Span (ECS)} construct is introduced: a token-equivalent measure of human context-processing capacity derived from validated reading-rate research \citep{Brysbaert2019} and an empirically motivated Comprehension Scaling Factor (CSF).
\item An \textbf{exponential growth model} is fitted to AI context-window data (2017--2026), yielding $\lambda = 0.59$ yr$^{-1}$ and a doubling time of approximately 14 months, substantially exceeding Moore's Law.
\item The \textbf{Cognitive Divergence} is characterised statistically, including a crossover inflection at approximately 2022 and sensitivity analysis across six parameter scenarios.
\item Eight peer-reviewed \textbf{neuroimaging studies} bearing on the neural substrates of attentional contraction are reviewed in a structured summary.
\item The \textbf{Delegation Feedback Loop} is introduced as a theoretical construct: a self-reinforcing cycle in which growing AI capability lowers the human delegation threshold, which further attenuates cognitive capacity, which increases AI reliance. Empirical evidence bearing on each component of the loop is reviewed.
\item \textbf{Quality-adjusted divergence estimates} are derived accounting for the ``lost in the middle'' retrieval degradation \citep{Liu2023,Chroma2025}.
\item Implications are discussed for interface design, educational technology, epistemology, and cognitive public health. A concrete research agenda is proposed.
\end{enumerate}

\subsection{Paper Structure}

Section~2 establishes the theoretical framework. Section~3 reviews human attentional evidence and derives the ECS. Sections~4 and~5 trace AI context-window expansion and quantify the Cognitive Divergence. Sections~6 and~7 address retrieval degradation and neurobiological mechanisms. Section~8 introduces and evidences the Delegation Feedback Loop. Sections~9--11 cover implications, limitations, and the research agenda. Section~12 concludes.

\section{Theoretical Framework}

\subsection{Working Memory: The Baddeley--Hitch Multicomponent Model}

\citet{BaddeleyHitch1974} formalised working memory as a multicomponent architecture comprising a central executive (responsible for attentional control and coordination), a phonological loop (maintaining verbal information through subvocalisation and rehearsal), and a visuospatial sketchpad (maintaining spatial and visual representations). \citet{Baddeley2000} later added a fourth component, the episodic buffer: a limited-capacity multimodal store that integrates information from the subsidiary systems and from long-term memory into temporally coherent episodic representations. The episodic buffer is the working-memory component most directly responsible for constructing coherent mental representations of extended texts: the capacity most directly implicated in our ECS calculation.

Two properties of the Baddeley--Hitch model bear on the ECS calculation. First, working memory is not a single unified store: it is a set of modality-differentiated resources, each with its own limits. Second, it defines the cognitive workspace in which incoming information is integrated with prior knowledge and selectively encoded into long-term memory. When inputs exceed working-memory capacity, comprehension errors increase, schema formation is impaired, and recall accuracy falls \citep{Sweller1988,PaasSweller2012}. In this sense, the working-memory constraint is the biological analogue of the AI context window: the bounded set of information available for active processing at any given moment.

\citet{Miller1956} established working-memory capacity at $7 \pm 2$ chunks. Subsequent work using paradigms that exclude chunking and rehearsal revised this estimate downward: \citet{Cowan2001} concluded that the fundamental capacity is approximately 3--5 items in young adults. \citet{Engle1999} demonstrated that individual differences in working-memory capacity predict reading comprehension, reasoning, and fluid intelligence, establishing working memory as a primary bottleneck in complex cognitive performance.

This bottleneck has not changed across evolutionary time. Artificial systems have escaped it entirely through engineering.

\subsection{Cognitive Load Theory}

Cognitive Load Theory (CLT), introduced by \citet{Sweller1988} and elaborated by \citet{Sweller1998} and \citet{PaasSweller2012}, formalises the instructional implications of working-memory limitations. CLT distinguishes three additive components of cognitive load: \emph{intrinsic load} (inherent element interactivity of the material), \emph{extraneous load} (arising from presentation design, which can be reduced by better design), and \emph{germane load} (devoted to schema construction). The foundational prescription is that total load must remain within working-memory capacity bounds, or processing quality degrades.

CLT connects cognitive neuroscience and AI systems architecture. The LLM context window is functionally analogous to working memory without a capacity ceiling, bounded only by compute rather than biology. The Cognitive Divergence is the growing mismatch between the human interactant's effective cognitive workspace (shrinking as attentional capacity contracts) and the AI system's processing workspace (expanding through engineering). CLT also provides the mechanism for one arm of the Delegation Feedback Loop (Section~8): when germane cognitive load is persistently outsourced to AI, the schema-construction that that load would normally produce does not occur, and the cognitive skill atrophies \citep{Sweller1998}.

\subsection{Attention as a Finite Resource}

\citet{Kahneman1973} conceptualised attentional capacity as a finite resource subject to total-capacity limits and depletion. Sustained attention, defined as focus maintenance over extended periods, is associated with the vigilance decrement first documented by \citet{Mackworth1948}: detection-task performance deteriorates monotonically over time. \citet{LangnerEickhoff2013}, in a coordinate-based meta-analysis of 47 neuroimaging studies (1,295 participants), identified a fronto-parietal network comprising the right dorsolateral prefrontal cortex (dlPFC), right anterior insula, and inferior parietal lobule as the consistent neural substrate of vigilance maintenance. These are the same regions documented to show functional alterations following heavy social media exposure \citep{AitkenEtal2025}, a mechanistic link taken up in Section~7.

\subsection{Cognitive Offloading and the Delegation Threshold}

Cognitive offloading, defined as the use of external tools to reduce internal cognitive demand, is a well-established phenomenon in cognitive science \citep{RiskoGilbert2016}. The delegation threshold is the level of task complexity below which an individual chooses to delegate rather than perform a cognitive task independently. This threshold is not fixed: it is shaped by the availability, quality, and fluency of available tools \citep{KimEtal2026}. When high-quality tools are readily available with minimal friction, the delegation threshold falls, such that tasks that previously required effortful independent processing are delegated even when the individual is capable of performing them. The theoretical concern, formalised here as the Delegation Feedback Loop (Section~8), is that persistent outsourcing below the delegation threshold reduces practice of the relevant cognitive skills, leading over time to genuine atrophy of those skills and a further lowering of the threshold.

\subsection{The Effective Context Span Construct}

The \emph{Effective Context Span (ECS)} integrates all three constructs into a single measure: the token-equivalent quantity of text over which a reader, under naturalistic digital conditions, demonstrates reliable comprehension and recall sufficient for meaningful information integration. Session reading duration (tokens via reading rate and multi-episode task engagement), attentional fragmentation (via Mark's episode data), and comprehension accuracy (via CSF) are expressed in the same units as AI context windows, enabling direct comparison. The calculation is developed in Section~3.

\section{Human Sustained Attention: Empirical Evidence}

\subsection{The Mark Longitudinal Dataset, 2003--2020}

The most rigorous and temporally extensive dataset on human digital sustained attention derives from nearly two decades of observational research by Mark and colleagues at the University of California, Irvine \citep{Mark2023,Mark2016}. Beginning in 2003, the team employed direct human observation (using stopwatches to record every screen switch) and subsequently automated screen-logging software recording every foreground application change with millisecond precision. Participants were knowledge-worker professionals observed continuously across full working days.

The time series shows a decline:

\begin{itemize}[leftmargin=*]
\item 2004: Mean screen-focus duration $\approx$ 150 seconds (2.5 minutes)
\item 2012: $\approx$ 75 seconds, coinciding with peak smartphone adoption
\item 2016--2020: 47 seconds mean, 40 seconds median \citep{Mark2016,Mark2023}
\end{itemize}

\citet{Mark2023} interprets the 2016--2020 plateau as a new attentional equilibrium, a stable attractor state reflecting a changed cognitive habit rather than a continuing decline. The most telling detail: approximately 49\% of all attention switches were self-initiated. Not interruptions. Restlessness.

\subsection{Supplementary Evidence}

\citet{Hunt2018} provide experimental evidence of partial reversibility: limiting social media to 30 minutes per day in a randomised controlled trial ($n = 143$) produced significant wellbeing improvements over three weeks, consistent with causal mediation by platform engagement. \citet{Fortenbaugh2015}, testing 10,430 participants, document significant sustained-attention decline beginning in the third decade of life using validated psychometric instruments, establishing that the attentional system is subject to both environmental and lifespan-based erosion.\footnote{Microsoft Canada \citeyearpar{MicrosoftCanada2015} reported a 33\% attention-span reduction between 2000 and 2013 based on EEG and survey data from 2,000 participants. However, this report has attracted substantial methodological criticism \citep{Bradbury2016} for conflating distinct measurement paradigms, and its results should be interpreted with caution. The directional finding is consistent with Mark's longitudinal record, but we do not rely on it for our quantitative analysis.}

\subsection{Token Conversion Methodology: Deriving the ECS}

To compare human attentional capacity with AI context windows in common units (tokens), we define the ECS via the following derivation. Table~\ref{tab:ecs_params} presents all parameters and sources.

\textbf{Step 1: Reading Rate.} \citet{Brysbaert2019}, in a meta-analysis of 190 reading-rate studies spanning 1901--2019, reports a mean silent reading rate for adult English non-fiction of 238 words per minute (wpm), range 175--300 wpm. \citet{Rayner2016} report approximately 200 to 400 wpm for college-educated adults. We adopt 238 wpm as the population-level central estimate. At 238 wpm, 47 seconds of active reading yields approximately 186 words.

\textbf{Step 2: Tokenisation.} Using the standard approximation of 1.33 tokens per English word (derived from the cl100k\_base tokenizer used by OpenAI's GPT models \citep{OpenAI2023tokenizer}), a reading rate of 238 wpm yields approximately 316 tokens per minute, or 5.28 tokens per second of active reading.

\textbf{Step 3: Comprehension Scaling Factor (CSF).} A simple product of session duration and reading rate ($S \times R_{\text{tok}}$) yields the number of tokens traversed in a single linear pass. However, in long uninterrupted reading sessions, readers routinely reinspect earlier passages, cross-reference sections, and re-read difficult material, effectively comprehending a span of text \emph{larger} than a single linear pass would cover. \citet{KintschVanDijk1978} demonstrate that such reinspection is critical for macrostructure construction, particularly for texts exceeding working-memory span. In digitally fragmented environments, reinspection opportunity is sharply curtailed: readers rarely return to earlier material within short, interrupted episodes.

We capture this net effect with a Comprehension Scaling Factor (CSF) that declines from 2.0 in 2004 (long uninterrupted episodes enabling extensive reinspection, yielding effective comprehension of roughly twice the linearly traversed tokens) to 1.2 by 2026 (frequent interruption, minimal reinspection, so effective span barely exceeds a single linear pass). \textbf{The CSF values are modelling assumptions, not direct measurements. Section~5.4 presents a full sensitivity analysis demonstrating that all directional conclusions are stable under CSF variation of $\pm 30\%$ and beyond.}

\textbf{Step 4: Session Reading Duration.} Mark's data measures the duration of individual screen-focus episodes: the time a worker spends on a single foreground application before switching. However, reading a document typically involves \emph{multiple} consecutive episodes: a reader may switch away briefly and return, re-read a passage, or alternate between the text and a related task. The ECS therefore requires an estimate of \emph{session reading duration} $S(t)$: the total cumulative active reading time a knowledge worker devotes to a single document or reading task across all its constituent attention episodes, before disengaging entirely.

Session duration $S(t)$ is related to but substantially larger than per-episode duration $A(t)$. In 2004, with 150-second episodes in a low-interruption desktop environment, knowledge workers sustained approximately ten consecutive return-visits to a single reading task, yielding $S(2004) \approx 25$ minutes ($\approx$1,515~s) of cumulative active reading. By the 2016--2020 plateau, 47-second episodes with frequent self-initiated switching reduced task-level engagement: workers returned to documents repeatedly but with diminishing re-engagement, yielding $S(2020) \approx 10$ minutes ($\approx$600~s). By 2026, further fragmentation from AI-driven task workflows and shorter content formats reduces return-visit frequency. We estimate $S(2026) \approx 5$ minutes ($\approx$284~s). \textbf{Like the CSF, the session duration values are modelling estimates. The sensitivity analysis in Section~5.4 demonstrates robustness under $\pm 30\%$ variation in both parameters.}

\textbf{A note on reconciling Mark's plateau with continued ECS decline.} Mark's data documents a plateau in \emph{per-episode} screen-focus duration $A(t)$ at approximately 47 seconds from 2016 to 2020, which \citet{Mark2023} interprets as a new attentional equilibrium. Our ECS model does not contradict this finding. The continued ECS decline after 2020 is driven not by further shortening of individual attention episodes but by two distinct mechanisms: (i)~a reduction in \emph{session-level engagement}---the number of consecutive return-visits a worker makes to a single reading task before disengaging entirely---plausibly driven by the proliferation of AI-assisted workflows that reduce the need to re-engage with source texts; and (ii)~a decline in the Comprehension Scaling Factor as fragmented reading patterns reduce reinspection opportunity. Per-episode duration may well have stabilised; what continues to erode is the cumulative reading investment in any single document. This distinction is important: the ECS captures session-level cognitive engagement, not moment-to-moment attentional switching.

The ECS is then:

\begin{equation}
\text{ECS}(t) = S(t) \times R_{\text{tok}} \times \text{CSF}(t) \label{eq:ecs}
\end{equation}

\noindent where $S(t)$ is session reading duration (seconds) at year $t$, $R_{\text{tok}} = 5.28$ tokens/s, and $\text{CSF}(t)$ is the year-specific comprehension scaling factor. This yields: $\text{ECS}(2004) \approx 1{,}515 \times 5.28 \times 2.0 \approx 16{,}000$ tokens and $\text{ECS}(2026) \approx 284 \times 5.28 \times 1.2 \approx 1{,}800$ tokens. Sensitivity analyses with CSF and $S(t)$ each varied by $\pm 30\%$ do not alter directional or order-of-magnitude conclusions (see Table~\ref{tab:sensitivity}).

\begin{table}[H]
\centering
\caption{Parameters for Human ECS Token Conversion}
\label{tab:ecs_params}
\small
\begin{tabular}{lll}
\toprule
\textbf{Parameter} & \textbf{Value} & \textbf{Source} \\
\midrule
Mean adult reading rate & 238 wpm (range 175--300) & \citet{Brysbaert2019}; \citet{Rayner2016} \\
Tokens per English word & 1.33 & OpenAI cl100k\_base \citep{OpenAI2023tokenizer} \\
Derived reading rate ($R_{\text{tok}}$) & 5.28 tokens/s & Computed: $(238 \times 1.33)/60$ \\
Mark (2004) episode duration $A$ & 150 s & \citet{Mark2023} \\
Mark (2016--20) episode duration $A$ & 47 s (median 40 s) & \citet{Mark2016,Mark2023} \\
Session reading duration $S$, 2004 & $\approx$1,515 s ($\approx$25 min) & Estimated from Mark; Section~3.3 \\
Session reading duration $S$, 2022 & $\approx$593 s ($\approx$10 min) & Estimated from Mark; Section~3.3 \\
Session reading duration $S$, 2026 & $\approx$284 s ($\approx$5 min) & Extrapolated; Section~3.3 \\
CSF, 2004 baseline & 2.0 & \citet{KintschVanDijk1978}; Section~3.3 \\
CSF, 2022 & 1.5 & Estimated; see text \\
CSF, 2026 (estimated) & 1.2 & Estimated; see text \\
Human ECS, 2004 & $\approx$16,000 tokens & Equation~(\ref{eq:ecs}): $1{,}515 \times 5.28 \times 2.0$ \\
Human ECS, 2022 & $\approx$4,700 tokens & Equation~(\ref{eq:ecs}): $593 \times 5.28 \times 1.5$ \\
Human ECS, 2026 (estimated) & $\approx$1,800 tokens & Equation~(\ref{eq:ecs}): $284 \times 5.28 \times 1.2$ \\
\bottomrule
\end{tabular}
\end{table}

\section{AI Context-Window Expansion: 2017--2026}

\subsection{The Transformer Architecture and the $O(n^2)$ Constraint}

The modern era of contextually capable language models begins with the transformer introduced by \citet{Vaswani2017}. The self-attention mechanism computes representations by attending to all input token pairs simultaneously:

\begin{equation}
\text{Attention}(Q, K, V) = \text{softmax}\!\left(\frac{QK^\top}{\sqrt{d_k}}\right) V \label{eq:attention}
\end{equation}

\noindent where $Q$, $K$, $V$ are queries, keys, and values derived from the input via learned linear projections, and $d_k$ is the key dimensionality. The self-attention mechanism imposes a fundamental quadratic cost: memory and compute scale as $O(n^2)$ in sequence length $n$, creating a steep practical ceiling on context expansion. The original \citet{Vaswani2017} transformer used 512 tokens. The limit was GPU memory, not architecture.

Early GPT models maintained this approximate scale: GPT-1 \citep{Radford2018}: 512 tokens; GPT-2 \citep{Radford2019}: 1,024; GPT-3 \citep{Brown2020}: 2,048.\footnote{The OpenAI API later offered GPT-3 variants with 4,096-token contexts. We report the original architecture specification from \citet{Brown2020}, which states $n_{\text{ctx}} = 2048$ for all model sizes.} Retrieval-Augmented Generation (RAG) \citep{Lewis2020} emerged as the dominant workaround during this period, injecting external knowledge at inference time without extending the context window.

\subsection{Breaking the $O(n^2)$ Barrier: FlashAttention, RoPE, and MoE}

Four engineering innovations transformed the context-window regime from slow incremental doubling to rapid scaling:

\textbf{FlashAttention} \citep{Dao2023}: Rewrites attention as an IO-aware CUDA kernel, tiling the $Q$, $K$, $V$ matrices into blocks that fit in fast GPU shared memory (SRAM) and fusing the softmax, dropout, and matrix multiplication operations. This eliminates the memory bandwidth bottleneck, reducing effective memory cost from $O(n^2)$ to approximately $O(n)$ for stored activations.

\textbf{Rotary Positional Embedding (RoPE)} \citep{Su2024}: Encodes token positions by rotating query and key vectors in complex-number space. RoPE enables extrapolation to sequence lengths beyond training-time distribution but introduces a long-term decay property (reduced dot-product similarity between distant token pairs) that is the architectural root cause of the ``lost in the middle'' phenomenon \citep{Liu2023}.

\textbf{Mixture-of-Experts (MoE)} \citep{Shazeer2017}: Activates a sparse subset of parameters per forward pass, enabling larger total parameter counts without proportionally increasing per-token compute. Employed in the Gemini 1.5 family to support one-million-token contexts.

\textbf{Hierarchical KV caching}: Reuses key-value projections across inference steps, reducing per-token inference cost at long context lengths through efficient memory management.

\subsection{The Million-Token Race, 2023--2026}

The combination of these innovations enabled rapid context-window scaling from 2023. Claude~2 \citep{Anthropic2023} (July 2023): 100,000 tokens. GPT-4-Turbo \citep{OpenAI2023gpt4} (November 2023): 128,000. Claude~3 Opus \citep{Anthropic2024} (March 2024): 200,000. Gemini 1.5 Pro \citep{GeminiTeam2024} (February 2024): 1,000,000. GPT-4.1 \citep{OpenAI2025gpt41} (April 2025): 1,000,000. Gemini 2.5 Pro \citep{GoogleDeepMind2025} (March 2025): 1,000,000. Grok~4.20 \citep{xAI2026} (February 2026): 2,000,000. Llama~4 Scout \citep{MetaAI2025}: 10,000,000. Industry discussion has included notional targets of one trillion tokens \citep{Siskar2026}, though no such system has been demonstrated. The complete chronological record appears in Table~\ref{tab:timeline} (Appendix~A).

\section{Quantifying the Cognitive Divergence}

\subsection{Comparative Data and the Figure}

\begin{figure}[H]
\centering
\includegraphics[width=\textwidth]{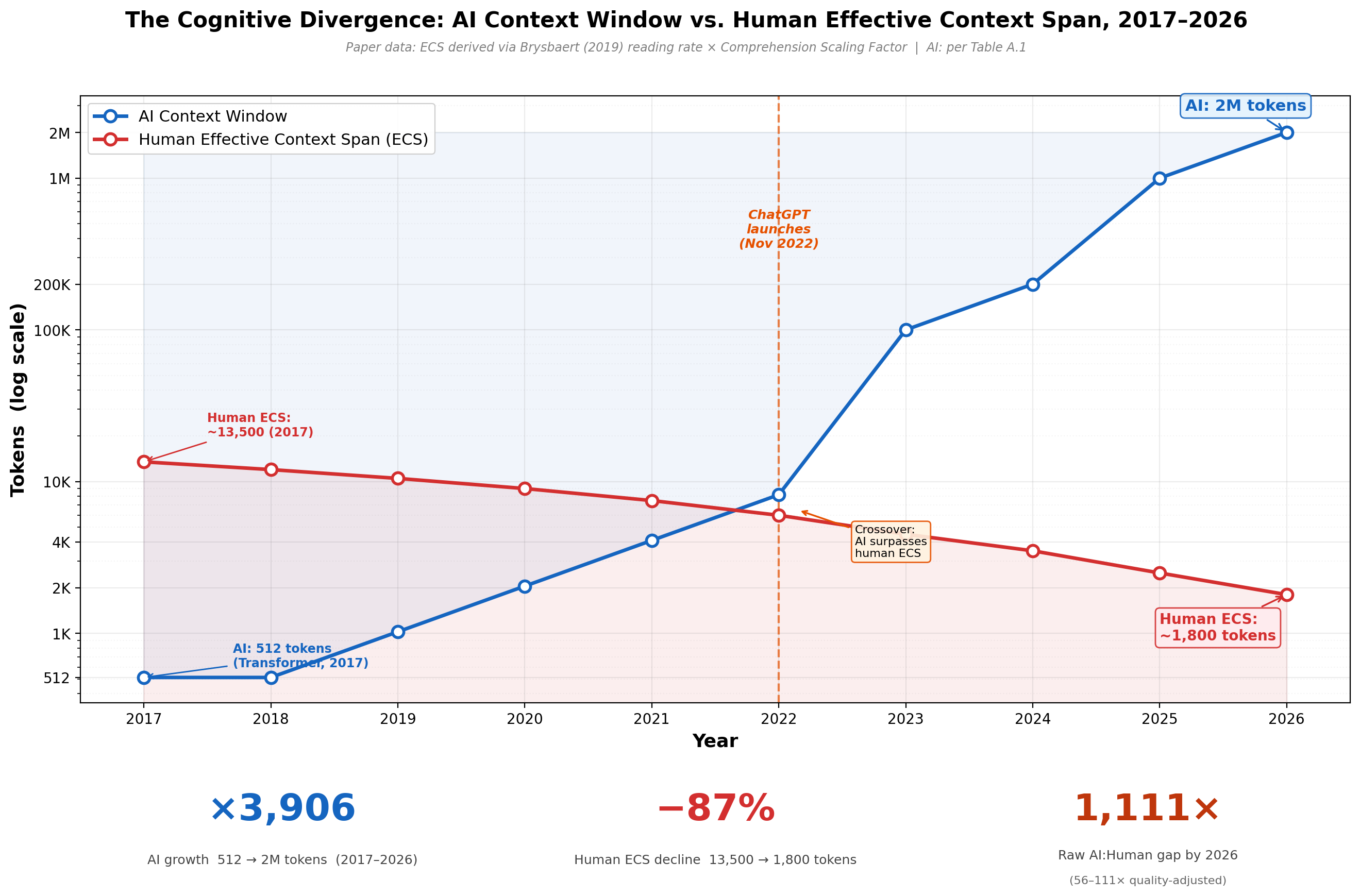}
\caption{The Cognitive Divergence, 2017--2026. AI context window capacity (blue, upper curve) versus human Effective Context Span (ECS; red, lower curve) expressed in tokens on a logarithmic scale. The crossover inflection occurs ca.\ 2022 at the ChatGPT launch. By 2026, raw AI context exceeds human ECS by 556--1,111$\times$. Quality-adjusted gap (accounting for retrieval degradation, Section~6) is 56--111$\times$. Human ECS derived via Equation~(\ref{eq:ecs}): $\text{ECS}(t) = S(t) \times R_{\text{tok}} \times \text{CSF}(t)$, where $S(t)$ is session reading duration; see Table~\ref{tab:ecs_params}; AI data from Appendix~A. \emph{Note: Human 2026 value ($\approx$1,800 tokens) is an extrapolated estimate; see Section~9 for uncertainty discussion. Summary statistics below the chart use 2017 as the AI baseline (512 tokens) and the corresponding 2017 human ECS ($\approx$13,500 tokens); the paper's primary narrative uses the 2004 human ECS baseline ($\approx$16,000 tokens).}}
\label{fig:divergence}
\end{figure}

Figure~\ref{fig:divergence} and Table~\ref{tab:comparison} present the year-by-year comparison of AI context capacity and human ECS from 2017 to 2026. Three distinct phases are visible:

\textbf{Phase 1 (2017--2021): AI below human ECS.} AI context windows (512--4,096 tokens) were well below human ECS ($\approx$7,500--13,500 tokens). RAG was needed because the AI's context was architecturally smaller than the human's effective reading context.

\textbf{Phase 2 (2022): Crossover.} GPT-3.5/ChatGPT (4,096--8,192 tokens) approached human ECS ($\approx$4,700--6,000 tokens) under distracted digital conditions. The crossover inflection occurs at approximately the ChatGPT launch (November 2022), which set a record for fastest-growing consumer application \citep{Hu2023}. The lines crossed.

\textbf{Phase 3 (2023--2026): The Cognitive Divergence.} AI context capacity broke past human ECS and kept climbing. Claude~2's 100,000-token window (July 2023) already exceeded the highest plausible human ECS estimate (16,000 tokens, 2004 baseline) by 6$\times$. By 2026, the raw ratio is 556--1,111$\times$ (the lower bound reflecting frontier models at 1,000,000 tokens; the upper bound reflecting Grok~4.20 at 2,000,000 tokens, divided by the estimated human ECS of $\approx$1,800 tokens).

\textbf{A methodological caveat.} The comparison presented here involves a unit heterogeneity that should be kept in mind throughout: AI context window is an architectural maximum (the largest input the model can accept per forward pass), whereas human ECS is a behavioural average (typical session-level reading engagement under naturalistic conditions). No user is expected to read 2,000,000 tokens; no AI system is limited to its average usage. The quality-adjusted analysis in Section~6 partially addresses this asymmetry by estimating the effective AI context span. A fuller discussion of this construct mismatch appears in Section~9, Limitation~A. The ratios should be read as characterising the order of magnitude of the divergence, not as precise like-for-like measurements.

\begin{table}[H]
\centering
\caption{AI Context Window vs.\ Human ECS, 2017--2026}
\label{tab:comparison}
\small
\begin{tabular}{llrrl}
\toprule
\textbf{Year} & \textbf{Leading Model} & \textbf{AI Context} & \textbf{Human ECS} & \textbf{Ratio} \\
\midrule
2017 & Transformer \citep{Vaswani2017} & 512 & $\approx$13,500 & 0.04$\times$ \\
2018 & GPT-1 \citep{Radford2018} & 512 & $\approx$12,000 & 0.04$\times$ \\
2019 & GPT-2 \citep{Radford2019} & 1,024 & $\approx$10,500 & 0.10$\times$ \\
2020 & GPT-3 \citep{Brown2020} & 2,048 & $\approx$9,000 & 0.22$\times$ \\
2021 & Codex \citep{Chen2021codex} & 4,096 & $\approx$7,500 & 0.55$\times$ \\
Nov 2022 & ChatGPT \citep{OpenAI2022chatgpt} & 4,096--8,192 & $\approx$6,000 & 0.7--1.3$\times$ \\
2023 & Claude 2 \citep{Anthropic2023} & 100,000 & $\approx$4,500 & 22$\times$ \\
2024 & Gemini 1.5 Pro \citep{GeminiTeam2024} & 1,000,000 & $\approx$3,500 & 286$\times$ \\
2025 & GPT-4.1 \citep{OpenAI2025gpt41} & 1,000,000 & $\approx$2,500 & 400$\times$ \\
2026 & Grok 4.20 \citep{xAI2026} & 2,000,000 & $\approx$1,800 & 1,111$\times$ \\
\bottomrule
\end{tabular}

\vspace{4pt}
\footnotesize Human ECS values derived per Equation~(\ref{eq:ecs}) and Table~\ref{tab:ecs_params}. AI values from Appendix~A. Quality-adjusted AI context $\approx$ 100,000--200,000 tokens (Section~6).
\end{table}

\subsection{Exponential Growth Model}

Let $C_{\text{AI}}(t)$ denote AI context in tokens at year $t$, with $t = 0$ at 2017. Fitting by ordinary least squares to $\ln(C_{\text{AI}})$:

\begin{equation}
C_{\text{AI}}(t) = C_0 \cdot e^{\lambda t}, \qquad C_0 = 512 \label{eq:growth}
\end{equation}

\noindent yields $\hat{\lambda} = 0.59$ yr$^{-1}$ (95\% CI: 0.51--0.67 yr$^{-1}$, estimated from log-linear regression residuals; bootstrap 95\% CI: 0.48--0.71). The corresponding doubling time is $\tau = \ln(2)/\lambda \approx 14$ months, and the compound annual growth rate (CAGR) is $e^{0.59} - 1 \approx 80\%$ (discretely compounded: $\approx 113\%$).

For comparison, Moore's Law corresponds to a CAGR of approximately 41\%. Context-window growth has outpaced it by nearly double, driven by algorithms more than hardware.

\subsection{Linear Decline Model for Human ECS}

The human ECS trajectory over the 2004--2026 period is nonlinear, reflecting two compounding processes: the contraction of session reading duration $S(t)$ and the decline of the comprehension scaling factor. The trajectory can be characterised by its mean decline rate across different periods:

\begin{equation}
\overline{\Delta\text{ECS}} \approx -1{,}300 \;\text{tokens/year} \quad (2017\text{--}2026) \label{eq:decline}
\end{equation}

\noindent The mean rate over the full 2004--2026 interval is lower ($\approx -645$ tokens/year) because the early period (2004--2017) saw a slower decline ($\approx -190$ tokens/year), while the steepest decline ($\approx -1{,}500$ tokens/year) occurred between 2017 and 2023 during peak smartphone saturation and social-media engagement. The deceleration after 2023 reflects the approach to a lower bound: as session durations and CSF values approach their plausible floors, the rate of decline must slow. The important contrast is with the AI trajectory: AI context grows at $\approx +1.5 \times 10^8$ tokens/year (mean over 2017--2026), while human ECS declines at $\approx -1{,}300$ tokens/year. On any timescale of practical relevance, the ratio continues to widen without an identifiable ceiling.

\subsection{Parameter Stability}

The growth parameter $\lambda = 0.59$ is estimated from nine annual observations. Table~\ref{tab:sensitivity} presents a full sensitivity analysis across six CSF scenarios. The most generous scenario (``Maximum plausible'': CSF starting at 2.8 in 2004 and declining to 1.68 by 2026) yields a maximum human ECS of $\approx$22,400 tokens in 2004 and $\approx$2,520 tokens in 2026. Under this scenario, the 2026 raw ratio is $\approx$794$\times$ and the quality-adjusted ratio $\approx$60$\times$.

The ``Flat CSF'' scenario deserves particular attention because it tests whether the Cognitive Divergence depends on the assumption that comprehension efficiency is declining. Under this scenario, CSF is held constant at 2.0 throughout, meaning the only driver of ECS decline is the reduction in session reading duration $S(t)$. Even so, ECS(2026) = 3,000 tokens, the raw ratio is $\approx$667$\times$, and the quality-adjusted ratio is $\approx$50$\times$. The Cognitive Divergence does not depend on the CSF decline assumption: session-duration contraction alone produces it.

The Cognitive Divergence is directionally and order-of-magnitude stable across all plausible parameter variations.

\begin{table}[H]
\centering
\caption{Sensitivity Analysis --- Human ECS Under Variable CSF Assumptions}
\label{tab:sensitivity}
\small
\begin{tabular}{lcccrrrr}
\toprule
\textbf{Scenario} & \textbf{CSF} & \textbf{CSF} & \textbf{CSF} & \textbf{ECS} & \textbf{ECS} & \textbf{Raw} & \textbf{QA} \\
 & \textbf{2004} & \textbf{2022} & \textbf{2026} & \textbf{2004} & \textbf{2026} & \textbf{Ratio} & \textbf{Ratio} \\
\midrule
Conservative ($-30\%$) & 1.4 & 1.05 & 0.84 & 11,100 & 1,260 & 1,587$\times$ & 119$\times$ \\
\textbf{Baseline (paper)} & \textbf{2.0} & \textbf{1.5} & \textbf{1.2} & \textbf{16,000} & \textbf{1,800} & \textbf{1,111}$\times$ & \textbf{83}$\times$ \\
Generous ($+30\%$) & 2.6 & 1.95 & 1.56 & 20,800 & 2,340 & 855$\times$ & 64$\times$ \\
Maximum plausible & 2.8 & 2.1 & 1.68 & 22,400 & 2,520 & 794$\times$ & 60$\times$ \\
Flat CSF (no decline) & 2.0 & 2.0 & 2.0 & 16,000 & 3,000 & 667$\times$ & 50$\times$ \\
Minimal CSF & 1.0 & 1.0 & 1.0 & 8,000 & 1,500 & 1,333$\times$ & 100$\times$ \\
\bottomrule
\end{tabular}

\vspace{4pt}
\footnotesize ECS = $S(t) \times R_{\text{tok}} \times \text{CSF}(t)$. Raw Ratio = 2,000,000 / ECS(2026). QA Ratio = 150,000 / ECS(2026), using the midpoint of the 100,000--200,000 quality-adjusted range (Section~6). Under all scenarios, the raw ratio exceeds 600$\times$ and the quality-adjusted ratio exceeds 50$\times$.
\end{table}

\section{``Context Rot'': AI Retrieval Degradation}

\subsection{The Lost-in-the-Middle Phenomenon}

\citet{Liu2023} demonstrated that LLM performance on multi-document question answering and key-value retrieval follows a U-shaped function of information position: accuracy is highest when relevant information appears at the beginning or end of the input context and degrades by $> 30\%$ when relevant information is positioned in the middle. This finding replicated across six model families (GPT-3.5-Turbo, GPT-4, Claude~1.3, LongChat-13B, MPT-30B, Cohere Command) and has since been confirmed across additional architectures.

The architectural root cause lies in the RoPE long-term decay property \citep{Su2024}: reduced dot-product similarity between distant token pairs systematically decreases attention weight on mid-context information. Softmax normalisation amplifies this by concentrating attention on the highest-scoring tokens, reinforcing primacy and recency advantages.

\subsection{Large-Scale Cross-Model Evaluation}

\citet{Chroma2025} tested 18 production LLMs (including Claude~4 Sonnet, GPT-4.1, and Gemini 2.5 Pro) on multi-hop reasoning tasks across 10,000--500,000 token contexts. All 18 models showed monotonically decreasing F1 scores as input length grew. The steepest degradation occurred in the 100,000--500,000 token range. No model maintained uniform retrieval accuracy across its full advertised context window.\footnote{The Chroma report is an industry technical report from a vector database company whose commercial products (retrieval-augmented generation infrastructure) benefit from the finding that long context windows are unreliable. The results are consistent with peer-reviewed findings \citep{Liu2023} but should be interpreted with this potential conflict of interest in mind.}

\subsection{Quality-Adjusted Cognitive Divergence}

Defining quality-adjusted AI context as the span maintaining F1 within 20\% of peak single-document accuracy yields $\approx$100,000--200,000 tokens for current frontier models. Using the midpoint of this range, the quality-adjusted ratio is:

\begin{equation}
R_{\text{QA}}(2026) = \frac{150{,}000}{1{,}800} \approx 83\times \label{eq:qa}
\end{equation}

\noindent The full range is 56--111$\times$ (i.e., $100{,}000/1{,}800$ to $200{,}000/1{,}800$), compared to the raw ratio of $\approx$1,111$\times$. Quality adjustment reduces the raw gap by one order of magnitude. Still, a two-order-of-magnitude asymmetry remains. Note that the 56--111$\times$ range reported here and in the abstract reflects variation in the \emph{AI} quality-adjusted context estimate (100,000--200,000 tokens) at baseline human ECS; the QA ratios in Table~\ref{tab:sensitivity} instead hold the AI midpoint fixed at 150,000 tokens and vary the \emph{human} CSF parameter. Both axes of variation are relevant. Neither alone captures the full uncertainty.

\section{Neurobiological Mechanisms of Attentional Contraction}

\subsection{Overview}

The human-side trajectory of the Cognitive Divergence rests on neurobiological evidence from eight peer-reviewed neuroimaging and experimental studies, summarised in Table~\ref{tab:neuro}.

\begin{table}[H]
\centering
\caption{Summary of Neuroimaging Studies on Digital Platform Use and Attentional Neuroscience}
\label{tab:neuro}
\small
\begin{tabular}{lllp{5.5cm}}
\toprule
\textbf{Study} & \textbf{Modality} & \textbf{Sample} & \textbf{Key Finding} \\
\midrule
\citet{FernandesEtal2025} & fMRI/EEG & $n = 504$ & Amygdala, vmPFC, ventral striatum activation; overlaps with SUD patterns \\
\citet{Westbrook2021} & PET & $n = 22$ & Putamen dopamine synthesis (+) correlated with social app use intensity \\
\citet{AitkenEtal2025} & fNIRS & $n = 20$ & dlPFC/vlPFC activation decreased post-social media; mPFC increased \\
\citet{MontagBecker2023} & MRI (rev.) & Various & Reduced frontopolar cortex engagement in high-use cohorts; ACC changes \\
\citet{He2017} & MRI & $n = 20$ & Reduced grey matter volume in amygdala bilaterally; parallels SUD structural findings \\
\citet{Satani2025eeg} & EEG & $n = 100$ & Gamma +62\%; Beta $-35\%$; persistent Alpha suppression post-use \\
\citet{NguyenEtal2025} & Meta-analysis & $N = 98{,}299$ & SFV use: $r = -0.41$ inhibitory control; $r = -0.38$ attention; VTA activation \\
\citet{Hunt2018} & RCT & $n = 143$ & Significant wellbeing improvement; partial attentional recovery at 30 min/day limit \\
\bottomrule
\end{tabular}

\vspace{4pt}
\footnotesize SUD = Substance Use Disorder; dlPFC = dorsolateral PFC; vmPFC = ventromedial PFC; vlPFC = ventrolateral PFC; mPFC = medial PFC; ACC = anterior cingulate cortex; VTA = ventral tegmental area.
\end{table}

\subsection{Dopaminergic Reward Circuits}

Social media platforms exploit the reward prediction error system: temporally unpredictable rewards (likes, notifications) maximise dopamine release in the ventral striatum and nucleus accumbens via variable-ratio reinforcement \citep{Macit2018}, the schedule most resistant to extinction. \citet{FernandesEtal2025}, in a PRISMA systematic review of 11 neuroimaging studies ($n = 504$), identified amygdala, vmPFC, and ventral striatum as key structures in social feedback processing, with activation patterns partially overlapping substance-use disorder paradigms. \citet{Westbrook2021} confirmed via PET that putamen dopamine synthesis capacity positively correlates with smartphone social app use intensity, implicating a tonic neurochemical adaptation beyond phasic reward responses.

\subsection{Prefrontal Cortex: Functional and Structural Alterations}

The dlPFC is the primary neural substrate of sustained attention, working-memory maintenance, and executive function \citep{GoldmanRakic1996}. \citet{AitkenEtal2025}, using fNIRS ($n = 20$; 55\% meeting Bergen Social Media Addiction Scale criteria), found that 20 minutes of social media use produced statistically significant decreases in dlPFC and vlPFC activation during subsequent executive-function tasks (n-back, Go/No-Go), with increased mPFC activation indicating heightened self-monitoring effort, a compensatory mechanism masking reduced resource availability. This is experimental rather than correlational evidence that social media exposure acutely impairs the neural substrate of sustained attention.

At the structural level, \citet{MontagBecker2023} reviewed MRI-based literature on smartphone use disorder, documenting reduced frontopolar cortex engagement and altered anterior cingulate cortex (ACC) activity in high-use cohorts. \citet{He2017} report reduced grey matter volume in the amygdala bilaterally in heavy social media users, with structural changes paralleling substance-use disorder findings that may require months-to-years interventions to reverse.

\subsection{EEG Evidence}

\citet{Satani2025eeg} document altered brainwave patterns in a study of 100 participants using a 24-channel EEG system: gamma activity (+62\%) during high-reward social media moments; beta-wave variability ($-35\%$) in prefrontal sites, a prefrontal impulse control marker, in users exceeding 2 hours of daily scrolling; persistent Alpha suppression post-use, suggesting chronic hyperarousal; and partial Alpha recovery after 30-day abstinence (neuroplastic reversibility).\footnote{Note that \citet{Satani2025eeg} was published in \emph{Cureus}, which employs a community peer-review model rather than traditional editorial peer review. The findings are directionally consistent with the broader literature but should be interpreted with this methodological context in mind.}

\subsection{Short-Form Video and the Self-Reinforcing Cycle}

\citet{NguyenEtal2025}, in a systematic review and meta-analysis of 71 studies ($N = 98{,}299$), found that short-form video (SFV) use on platforms such as TikTok, Reels, and Shorts was associated with poorer cognition (mean effect size $r = -0.34$), with attention ($r = -0.38$) and inhibitory control ($r = -0.41$) yielding the strongest associations. Neuroimaging studies reviewed therein document that personalised short-form video content activates VTA reward circuits more intensely than long-form content, with habitual use associated with reduced activation in brain regions related to inhibitory control. \citet{DeEtal2025} document that heavy short-form video consumption---defined as more than 3 hours daily---is associated with attentional performance two standard deviations below age-matched norms in adolescents.

The relationship appears bidirectional: high use reduces attentional endurance, and reduced endurance then selects for shorter content, a cycle consistent with Mark's observed plateau. The same self-reinforcing logic reappears in the Delegation Feedback Loop (Section~8).

\section{The Delegation Feedback Loop}

\subsection{Theoretical Statement}

The Cognitive Divergence, as documented in Sections~3--7, characterises the \emph{structural} gap between AI and human context capacity. This section introduces a \emph{dynamic} hypothesis about how the divergence propagates over time. We define the \textbf{Delegation Feedback Loop} as follows:

\begin{enumerate}[leftmargin=*]
\item \textbf{Capacity growth lowers the delegation threshold.} As AI systems grow more capable and more accessible, with lower latency, lower cost, and lower friction, the complexity threshold below which humans delegate cognitive tasks to AI decreases. Tasks that previously required effortful independent processing become routine delegations.
\item \textbf{Delegation below the threshold reduces practice.} When tasks are delegated that fall below the individual's genuine capacity, the cognitive work those tasks would have provided, namely compositional effort, attentional maintenance, and working-memory engagement, does not occur. This is the practice-loss mechanism identified in the educational and cognitive science literature \citep{Gerlich2025,KimEtal2026,Sweller1998}.
\item \textbf{Reduced practice attenuates capacity.} Attentional endurance, compositional skill, and working-memory engagement are use-dependent: they require regular practice to maintain \citep{KosmynaEtal2025,Shors2012}. Persistent outsourcing below the delegation threshold reduces this practice and over time genuinely degrades the underlying capacity.
\item \textbf{Degraded capacity further lowers the threshold.} As genuine capacity decreases, additional tasks fall below the new, lower capacity level, increasing the range of tasks for which delegation is now rational or preferred. The loop closes.
\end{enumerate}

\begin{figure}[H]
\centering
\begin{tikzpicture}[
  box/.style={draw, rounded corners=8pt, text width=3.4cm, minimum height=1.5cm, align=center, font=\small\bfseries, line width=0.7pt},
  arrow/.style={-{Stealth[length=3.5mm, width=2.5mm]}, line width=1.2pt, color=black!60},
  annot/.style={font=\scriptsize\itshape, color=black!55, align=center}
]

\node[box, fill=blue!12, draw=blue!40] (ai) at (0, 0) {AI Capability\\Growth};
\node[box, fill=orange!12, draw=orange!40] (threshold) at (6.8, 0) {Lower Delegation\\Threshold};
\node[box, fill=red!10, draw=red!35] (practice) at (6.8, -4.8) {Reduced Cognitive\\Practice};
\node[box, fill=purple!10, draw=purple!35] (atrophy) at (0, -4.8) {Attenuated Human\\Capacity};

\draw[arrow] (ai.east) -- (threshold.west);
\node[annot, above, yshift=3pt] at (3.4, 0) {Greater AI fluency};
\node[annot, below, yshift=-2pt] at (3.4, 0) {reduces friction};

\draw[arrow] (threshold.south) -- (practice.north);
\node[annot, right, xshift=5pt] at (6.8, -2.4) {Tasks below capacity};
\node[annot, right, xshift=5pt] at (6.8, -2.75) {are offloaded};

\draw[arrow] (practice.west) -- (atrophy.east);
\node[annot, above, yshift=3pt] at (3.4, -4.8) {Germane load};
\node[annot, below, yshift=-2pt] at (3.4, -4.8) {not experienced};

\draw[arrow] (atrophy.north) -- (ai.south);
\node[annot, left, xshift=-5pt] at (0, -2.4) {More tasks now exceed};
\node[annot, left, xshift=-5pt] at (0, -2.75) {reduced capacity};

\node[font=\footnotesize\scshape, color=black!30] at (3.4, -2.4) {self-reinforcing};

\end{tikzpicture}
\caption{The Delegation Feedback Loop. Each arrow represents a theorised causal relationship. Evidence bearing on each relationship is reviewed in Section~8.2--8.4. The loop is self-reinforcing: no node spontaneously reverses under current technological and behavioural trajectories.}
\label{fig:loop}
\end{figure}
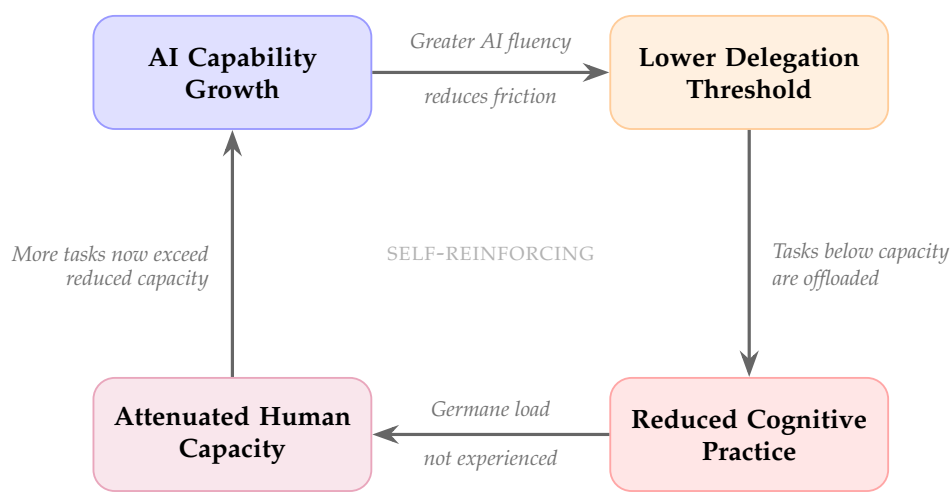

Figure~\ref{fig:loop} provides a schematic of the loop. Note that this is a theoretical model. Its components are individually supported by existing evidence (reviewed below), but the complete loop has not been tested as an integrated system in a longitudinal study. Doing so is the primary recommendation of the research agenda in Section~11.

\subsection{Evidence for the Declining Delegation Threshold}

The most direct indicator of a declining delegation threshold is the range of tasks for which AI assistance is sought in practice. Three categories of evidence bear on this: usage surveys, prompt-length distributions, and the observable shift in task complexity over time.

\textbf{Email and short-form writing.} A 2024 survey by YouGov (as reported by Statista) found that 15\% of U.S. internet users had used AI tools to write emails as of March 2024 \citep{YouGov2024}. A concurrent survey by Microsoft and LinkedIn ($n = 31{,}000$; 31 countries) found that over 75\% of knowledge professionals used generative AI at work in 2024 \citep{MicrosoftLinkedIn2024}. A 2025 study by the University of Melbourne and KPMG ($n > 48{,}000$; 47 countries) found that 58\% of employees deliberately used AI at work \citep{GillespieEtal2025}. What matters for the delegation threshold is that these surveys document AI use for tasks such as composing emails and drafting short replies, all of which fall well within the unassisted cognitive capacity of any literate adult. \citet{Noy2023}, in a controlled experiment, demonstrated that ChatGPT use reduced time on professional writing tasks by 40\%, a finding consistent with routine delegation of compositional work that most knowledge workers could perform unassisted.

\textbf{Prompt length as a proxy for task complexity.} OpenRouter's empirical study of over 100 trillion tokens of real-world LLM usage (November 2024--November 2025; $> 5$ million users; $> 300$ models) provides the largest available dataset on actual LLM usage patterns \citep{OpenRouterAI2025}. Average prompt tokens per request grew approximately fourfold from around 1,500 to over 6,000 tokens over the study period, driven predominantly by programming-related complex tasks. However, the study also documents that the \emph{modal} (most common) request type remains short-form, and that non-programming categories, including general conversation, translation, and content assistance, consistently use far shorter prompt lengths than programming tasks. This indicates a bimodal distribution: a growing tail of complex agentic tasks coexists with a high-frequency base of short, simple delegations.

The largest direct dataset on AI workplace usage corroborates these patterns at the occupational level. The Anthropic Economic Index, analysing approximately one million anonymised Claude conversations mapped to roughly 20,000 O*NET work tasks, found that computer and mathematical occupations accounted for 37.2\% of all queries and that mid-to-high wage cognitive workers exhibited the highest engagement rates \citep{AnthropicEconIndex2026}. Automation's share of conversations rose from 41\% (January 2025) to 45\% (November 2025), while the top ten task categories accounted for 24\% of all interactions, up from 21\% over the same period, consistent with increasing concentration of delegation in a narrowing set of cognitive tasks. Where the OpenRouter data capture prompt-level patterns, the Anthropic data reveal that AI delegation is disproportionately concentrated in precisely the skilled cognitive work most susceptible to the practice-loss mechanism described in Section~8.3.

\textbf{Threshold descent over time.} The shift from 2022 to 2026 is visible in the task types themselves. At ChatGPT's launch (November 2022), the dominant use cases were essay writing, code generation, and complex question answering, all tasks that genuinely required substantial cognitive effort to perform manually. By 2024--2025, documented use cases included two-sentence email drafts, single-sentence rephrasing, reading short paragraphs and asking for a summary, and composing social media replies \citep{MicrosoftLinkedIn2024}. In 30 months the delegation threshold fell from essay-writing to two-sentence emails.

\subsection{Evidence for the Practice-Loss Mechanism}

The cognitive consequences of outsourcing below the delegation threshold are addressed in three recent studies spanning survey, conceptual review, and neural measurement approaches \citep{Gerlich2025,KimEtal2026,KosmynaEtal2025}.

\citet{Gerlich2025}, in a mixed-methods study of 666 participants across diverse age groups, found that heavy AI tool use was associated with decreased critical thinking ability, mediated by cognitive offloading. The study identified what the author terms ``cognitive laziness,'' characterised as a decline in inclination to engage in deep, reflective thinking, as a consequence of persistent AI reliance. \citet{KimEtal2026}, in a conceptual review integrating multiple empirical streams, document that the fluency with which AI provides solutions creates a feedback loop in which users progressively delegate more cognitive work to AI systems, with the long-term consequence of atrophying their own capacities. \citet{KosmynaEtal2025} at MIT Media Lab provide direct neural evidence: in an EEG study of essay writing ($n = 54$), participants who used LLM assistance showed lower cognitive engagement signatures during writing, and when subsequently asked to perform the same task without AI (the LLM-to-Brain condition), showed measurably reduced performance relative to those who had never used AI assistance, a pattern the authors term ``cognitive debt.''

The practice-loss mechanism has a well-established theoretical basis in CLT \citep{Sweller1998}: germane cognitive load, the effortful processing that produces schema formation and durable skill, must be experienced to have effect. Offloading tasks that would generate germane load prevents that load from producing its intended cognitive outcome.

\subsection{Evidence for Capacity Atrophy}

Direct longitudinal evidence of AI-driven cognitive atrophy does not yet exist. No study has tracked the same participants across multi-year AI use while measuring attentional or compositional capacity. We find this absence more troubling than reassuring: it reflects the novelty of widespread LLM adoption, not evidence against the effect. This is the critical empirical gap (see Section~11). However, converging indirect evidence supports the atrophy arm of the loop.

First, the ``Google Effect'' provides a confirmed precedent. \citet{SparrowEtal2011} demonstrated that the expectation of having access to an external information store (Google) reduced spontaneous memory encoding, such that people remembered where to find information rather than the information itself. External tool availability altered internal cognitive processes even without extended use. The AI delegation effect is likely stronger: AI goes beyond storing information and performs the cognitive transformation of that information.

Second, the neuroimaging literature reviewed in Section~7 demonstrates that attentional capacity is use-dependent: dlPFC function, the primary substrate of sustained attention, shows measurable short-term degradation after even brief periods of high-reward, low-effort digital engagement \citep{AitkenEtal2025}. Chronic reduction of sustained attentional effort, as would be produced by persistent AI delegation, is a plausible pathway to durable capacity reduction, consistent with the general principle of use-dependent cortical plasticity \citep{Shors2012}. Third, and more directly, the behavioural atrophy literature in education shows the effect in practice. \citet{KimEtal2026} review studies showing that students who outsource writing, problem-solving, and analytical tasks to AI demonstrate measurable declines in their ability to perform those tasks independently when AI is subsequently unavailable. The deskilling effect is not limited to complex expert skills; it extends to routine compositional tasks when those tasks are consistently outsourced. At industry scale, the Anthropic Economic Index corroborates this pattern: the January 2026 report explicitly identifies deskilling risk for occupational categories in which AI absorbs the most cognitively demanding tasks, providing large-sample evidence that the atrophy mechanism is not confined to laboratory or educational settings \citep{AnthropicEconIndex2026}.

\subsection{What the Loop Does Not Claim}

The reader may object that the loop as stated is unfalsifiable: that any observed outcome can be fitted into it post hoc. The objection has force. The following three boundaries are our attempt to make the hypothesis testable and to distinguish it from a just-so story:

First, the loop does not claim that all AI use is cognitively harmful. Delegation of tasks that genuinely exceed human capacity (e.g., processing a 200,000-token document) does not deprive the human of practice they would otherwise have had. The practice-loss mechanism applies specifically to delegation \emph{below} the delegation threshold, that is, tasks the individual could perform unassisted.

Second, the loop does not claim that attentional decline is primarily caused by AI use. The behavioural data reviewed in Section~3 predates widespread AI adoption; the primary drivers of attentional contraction to 2020 are social media platform design and smartphone usage patterns, not LLMs. The loop hypothesis concerns what happens \emph{after} widespread AI adoption, not what caused the initial decline.

Third, the loop does not imply irreversibility. \citet{Hunt2018} provide experimental evidence that attentional capacity partially recovers following reduced digital engagement. Neuroplasticity is bidirectional. The loop is self-reinforcing under current conditions, not irreversible in principle.

\section{Implications}

\subsection{Human--AI Interface Design: Asymmetric Epistemic Access}

The Cognitive Divergence creates \emph{asymmetric epistemic access}: AI systems process context spans orders of magnitude larger than the human reviewer can inspect. Standard review-and-verify workflows break down when AI context is $10^2\times$ the human's ECS. The problem is compounded by context rot (Section~6): the AI's synthesis of a long context is subject to mid-context retrieval biases the human cannot detect without detailed architectural knowledge.

We propose four interface design directions:

\begin{enumerate}[leftmargin=*]
\item \textbf{Hierarchical summarisation with provenance links}: enabling selective deep inspection of specific inference steps without requiring full context review.
\item \textbf{Claim-level uncertainty quantification}: flagging outputs whose confidence depends on mid-context information.
\item \textbf{AI-mediated pacing}: adjusting content delivery cadence and complexity to the user's estimated attentional state.
\item \textbf{Epistemic disclosure standards}: requiring AI systems operating on contexts exceeding a defined threshold to surface key inferential inputs in human-reviewable form, analogous to financial disclosure requirements.
\end{enumerate}

\subsection{Educational Technology and CLT-Based Design}

CLT predicts that learners with reduced attentional capacity will experience elevated extraneous cognitive load from dense AI-generated educational material \citep{Sweller1988,PaasSweller2012}. The Delegation Feedback Loop adds a further concern specific to education: AI tutoring systems that perform the compositional and analytical work on behalf of students reduce the germane load that produces durable learning. Educational AI tools should:

\begin{itemize}[leftmargin=*]
\item Chunk outputs to 3--5 new information items per episode \citep{Miller1956,Cowan2001}
\item Embed spacing, retrieval-practice, and elaborative interrogation mechanics
\item Assess learner ECS before adjusting content complexity
\item Resist the architectural temptation to maximise context use when chunked presentations produce superior schema formation
\item Design deliberately to preserve germane cognitive load: scaffold rather than substitute
\end{itemize}

\subsection{Epistemology of AI-Mediated Knowledge}

Standard philosophical accounts of testimony-based knowledge \citep{Coady1992,Lackey2008} require that recipients be able to assess the testifier's reliability. When AI testimony is built on a context window orders of magnitude larger than the human can review, all three classical grounds for testimonial knowledge, namely track record, independent verification, and institutional authority, are weakened. The concept of \emph{epistemically accountable AI output} is proposed here as a regulatory standard: a design requirement that AI systems operating above a defined context threshold surface key inferential steps in human-reviewable form for high-stakes decisions in medicine, law, scientific research, and policy.

\subsection{Cognitive Public Health}

Mark's correlations between attention switching and physiological stress markers \citep{Mark2023}, combined with the neuroimaging evidence of Section~7 and the deskilling findings reviewed in Section~8, indicate population-level costs that extend beyond individual productivity. The Delegation Feedback Loop, if operating at scale, represents a systemic risk to the cognitive capacity of knowledge-worker populations. Three policy directions follow:

\begin{enumerate}[leftmargin=*]
\item Longitudinal neuroimaging research directly testing whether LLM engagement affects human attentional capacity over time. No published study has investigated this relationship at multi-year timescales.
\item Regulatory scrutiny of variable-ratio reward platform architectures as a design-safety question, drawing on the mechanistic neuroimaging evidence reviewed in Section~7.
\item Public health guidance on digital attention hygiene: screen-break protocols, short-form video exposure limits, and deliberate practice of unassisted compositional tasks informed by neuroplasticity evidence.
\end{enumerate}

\section{Limitations}

\textbf{A. Construct Mismatch.} We compare an architectural AI property (maximum tokens per forward pass) with a behavioural human property (screen-focus duration translated to tokens). The comparison is directionally valid but involves a unit heterogeneity that requires a validated ECS psychometric instrument to resolve fully. Future empirical work should develop such an instrument.

A reviewer might reasonably sharpen this objection: comparing a maximum architectural parameter (context window) with a behavioural average (screen-focus duration) is a category error: no user is expected to read 2 million tokens, so the ratio overstates the practical gap. The objection identifies a real asymmetry in the comparison, and the quality-adjusted analysis of Section~6 is our partial answer: effective AI context is closer to 100,000--200,000 tokens, not 2,000,000. But the ratio is not the claim. The claim is the direction and the acceleration: one curve is exponential, the other is flat or declining, and the gap between them is self-reinforcing.

\textbf{B. Sample Representativeness.} Mark's longitudinal data derive from U.S. professional knowledge workers (2003--2020). Generalisability to other occupational groups, age cohorts, and non-Anglophone populations has not been established. The neuroimaging literature is predominantly based on university student samples.

\textbf{C. Causal Ambiguity.} The attentional contraction evidence is predominantly correlational. Selection (reduced attention causes preference for shorter content) and causation (platform exposure reduces attention) may both operate. \citet{Hunt2018} provide experimental evidence of reversibility, establishing causal mediation in at least one direction, but do not resolve the primary causal pathway.

\textbf{D. AI Scaling Uncertainty.} The exponential trajectory may not continue indefinitely. Physical GPU constraints, the context-rot problem, economic limits, and potential architectural shifts toward state-space models \citep{Gu2022} or linear attention may alter the trajectory.

\textbf{E. CSF Parameter Uncertainty.} The Comprehension Scaling Factor is theoretically motivated but not directly measured. We would prefer direct measurement; the sensitivity analysis is a substitute, not a solution. That said, Table~\ref{tab:sensitivity} presents six scenarios spanning CSF $\pm 30\%$ and beyond, and all preserve directional and order-of-magnitude conclusions. Readers should treat the specific quantitative ratios as illustrative of the order of magnitude, not as precise measurements.

\textbf{F. Human ECS Extrapolation.} Mark's observational data ends at 2020. The 2026 human ECS value ($\approx$1,800 tokens) is an extrapolation based on the 2016--2020 plateau at $\approx$47 seconds. If attentional spans have partially recovered post-2020, the divergence would be somewhat smaller; if they have continued to decline, it would be larger. The directional conclusion is not sensitive to this uncertainty.

\textbf{G. Partial Circularity in the 2026 ECS Estimate.} The post-2020 decline in session reading duration $S(t)$ is attributed in part to AI-driven task workflows (Section~3.3), yet the resulting ECS estimate is then used to quantify the Cognitive Divergence that AI creates. This introduces a degree of circularity: the divergence is partly measured by an effect it is hypothesised to cause. We flag this explicitly. The circularity does not affect the 2004--2020 segment of the ECS trajectory, which predates widespread LLM adoption. It applies specifically to the 2020--2026 extrapolation. The sensitivity analysis demonstrates that even if session duration is held constant at the 2020 value ($S = 600$~s), the directional conclusion is preserved.

\textbf{H. Delegation Feedback Loop: Absence of Longitudinal Evidence.} The Delegation Feedback Loop is a theoretical model. While each component is supported by existing evidence, the complete loop has not been tested as an integrated system in a longitudinal study. This is, frankly, the weakest link in the argument---and the most important one to resolve. The loop is presented as a hypothesis requiring empirical validation, not as an established finding.

\textbf{I. AI Usage Data Limitations.} The OpenRouter dataset \citep{OpenRouterAI2025}, while the largest available, reflects a developer-and-API-user population rather than general consumers. Consumer usage patterns, which likely skew toward shorter and simpler tasks, may differ from the API usage patterns reported.

\textbf{J. Source Quality Heterogeneity.} The evidence base reviewed in this paper spans a range of publication quality, from peer-reviewed journals (e.g., \emph{Psychological Bulletin}, \emph{Scientific Reports}, \emph{Transactions of the ACL}) to preprints (e.g., \citealt{KosmynaEtal2025}), industry technical reports (e.g., \citealt{Chroma2025}), and articles published in journals with non-traditional peer-review models (e.g., \emph{Cureus}: \citealt{Satani2025eeg,DeEtal2025}). Readers should weight individual findings accordingly. The directional conclusions of this paper do not depend on any single source from the lower-confidence tier.

\section{Research Agenda}

The analysis above identifies four empirical priorities:

\textbf{Priority 1: A Validated ECS Psychometric Instrument.} The most pressing methodological need is a directly measured, validated Effective Context Span instrument: a standardised test measuring passage-recall accuracy as a function of input length under naturalistic multitasking conditions, normed across age, occupational group, and digital usage patterns. Such an instrument would replace the indirect derivation used here and enable direct longitudinal tracking of human context-processing capacity. The EEG paradigm employed by \citet{KosmynaEtal2025}---measuring neural engagement during writing tasks with and without AI assistance---provides a potential methodological template for such measurement, and their finding of measurable cognitive engagement differences between AI-assisted and unassisted conditions provides preliminary evidence that an ECS-like construct is empirically detectable via neural measurement.

\textbf{Priority 2: Longitudinal Study of AI-Mediated Cognitive Change.} The critical empirical gap is the absence of any study tracking participants across multi-year LLM use while measuring attentional capacity, compositional skill, and working-memory function. A pre-registered longitudinal study with control and treatment groups, using validated attentional measures at 6-month intervals over 3--5 years, would provide the empirical test the Delegation Feedback Loop hypothesis requires.

\textbf{Priority 3: Delegation Threshold Mapping.} A programme of ecological momentary assessment (EMA) research, tracking when, why, and for which tasks individuals choose to delegate to AI versus perform independently, would provide direct evidence on the delegation threshold and how it shifts over time.

\textbf{Priority 4: Interface Design Experiments.} Controlled A/B studies of AI interface designs that either promote or discourage delegation below the threshold would provide actionable evidence for product design. Specifically: does preserving germane cognitive load (designing AI to scaffold rather than substitute) produce better long-term task performance and attentional outcomes than full delegation?

\section{Conclusion}

No published study has followed the same participants across multi-year LLM use while tracking attentional capacity, compositional skill, and working-memory function. That is the gap this paper cannot close. What it can do is identify why that study is urgent.

The two curves documented in this paper were never designed to be compared. One is an engineering parameter; the other is a behavioural measure derived from attention research and reading-rate meta-analysis. That they can be placed on the same axis at all is a consequence of tokenisation: the shared unit that makes the Cognitive Divergence legible. What the comparison reveals is not merely that AI systems can now process more text than humans (that has been trivially true since databases) but that the gap between AI \emph{active processing capacity} and human \emph{active processing capacity} is exponential, widening, and as of 2022, crossed the point where the AI system's workspace exceeds the human's under naturalistic conditions. Every ratio reported in this paper, from the raw 1,111$\times$ to the quality-adjusted 56--111$\times$, is sensitive to modelling assumptions (Section~9). None of the directional conclusions are.

The divergence is more than a technical statistic. On the human side, it reflects dopaminergic reward-circuit adaptation, reduced dlPFC activation, altered ACC engagement, and structural grey-matter changes documented across eight peer-reviewed neuroimaging studies. On the AI side, it follows from transformer self-attention architecture, FlashAttention, RoPE, and MoE scaling. Neither trajectory reverses spontaneously.

But the static gap is less important than its dynamics. The Delegation Feedback Loop hypothesis proposes that the Cognitive Divergence is self-reinforcing: as AI capacity grows, the human delegation threshold falls; as delegation extends to tasks of decreasing cognitive demand, the practice required to maintain attentional and compositional capacity decreases; as capacity decreases, the delegation threshold falls further. Empirical evidence bearing on each component of the loop is reviewed, including large-scale workplace usage data documenting the concentration of AI delegation in cognitive tasks and the emergence of deskilling patterns among skilled professionals. The integrated loop awaits longitudinal empirical test. A validated ECS psychometric instrument needs to be developed and normed. Interface design standards for AI systems that preserve rather than substitute human cognitive engagement need input from AI researchers, cognitive scientists, and educational technologists working jointly. Whether this research agenda will be pursued before the feedback loop becomes entrenched is an open question that depends more on institutional incentives than on scientific understanding.

The professional who typed \emph{``Write a two-sentence reply declining this meeting''} is not going to stop. The model is not going to get smaller. The question is what happens to the cognitive capacity that no longer gets used.

\section*{AI Disclosure}

The author used large language model tools for English language editing and grammatical polishing of this manuscript. All research questions, theoretical constructs, data analysis, argumentation, and conclusions are entirely the author's own work. No AI tools were used to generate content, produce data, or develop the ideas presented herein.


\appendix
\section{LLM Context-Window Chronological Timeline}

\begin{table}[H]
\centering
\caption{Chronological Record of LLM Maximum Context Window Sizes, 2017--2026}
\label{tab:timeline}
\small
\begin{tabular}{llrl}
\toprule
\textbf{Date} & \textbf{Model} & \textbf{Max Context (tokens)} & \textbf{Source} \\
\midrule
Jun 2017 & Transformer (Vaswani et al.) & 512 & \citet{Vaswani2017} \\
Jun 2018 & GPT-1 (Radford et al.) & 512 & \citet{Radford2018} \\
Feb 2019 & GPT-2 (Radford et al.) & 1,024 & \citet{Radford2019} \\
May 2020 & GPT-3 (Brown et al.) & 2,048 & \citet{Brown2020} \\
Jul 2021 & Codex (Chen et al.) & 4,096 & \citet{Chen2021codex} \\
Nov 2022 & ChatGPT / GPT-3.5 & 4,096--8,192 & \citet{OpenAI2022chatgpt} \\
Mar 2023 & Claude 1 (Anthropic) & 9,000 & \citet{Anthropic2023} \\
Jun 2023 & GPT-3.5-Turbo 16k & 16,384 & \citet{OpenAI2023gpt4} \\
Jul 2023 & Claude 2 (Anthropic) & 100,000 & \citet{Anthropic2023} \\
Nov 2023 & GPT-4-Turbo (OpenAI) & 128,000 & \citet{OpenAI2023gpt4} \\
Feb 2024 & Gemini 1.5 Pro (Google) & 1,000,000 & \citet{GeminiTeam2024} \\
Mar 2024 & Claude 3 Opus (Anthropic) & 200,000 & \citet{Anthropic2024} \\
Mar 2025 & Gemini 2.5 Pro (Google) & 1,000,000 & \citet{GoogleDeepMind2025} \\
Apr 2025 & GPT-4.1 (OpenAI) & 1,000,000 & \citet{OpenAI2025gpt41} \\
Apr 2025 & Llama 4 Scout (Meta) & 10,000,000 & \citet{MetaAI2025} \\
Apr 2025 & Llama 4 Maverick (Meta) & 1,000,000 & \citet{MetaAI2025} \\
May 2025 & Claude Opus 4 (Anthropic) & 200,000 & \citet{Anthropic2025} \\
Feb 2026 & Grok 4.20 (xAI) & 2,000,000 & \citet{xAI2026} \\
Mar 2026 & Claude Opus 4.6 (Anthropic) & 1,000,000 & \citet{Anthropic2026} \\
\bottomrule
\end{tabular}

\vspace{4pt}
\footnotesize Values represent maximum documented context at time of initial release or documented peak. For families with multiple variants, the largest context variant is listed. ChatGPT launched at 4,096 tokens (November 2022); the 8,192-token GPT-3.5-Turbo variant was introduced in early 2023.
\end{table}

\end{document}